\documentclass{article}

\usepackage{spconf,amsmath,epsfig,amssymb}
\usepackage{capt-of}
\usepackage{multirow}
\usepackage{booktabs,adjustbox,graphicx}
\usepackage{enumitem}
\usepackage{color}
\usepackage[colorlinks,
            linkcolor=red,
            anchorcolor=blue,
            citecolor=green]{hyperref}
	
\usepackage[capitalise]{cleveref}
\Crefname{equation}{Eq.}{Eqs.}
\Crefname{figure}{Fig.}{Figs.}
\Crefname{table}{Tab.}{Tabs.}

\usepackage{placeins}

\let\OLDthebibliography\thebibliography
\renewcommand\thebibliography[1]{
  \OLDthebibliography{#1}
  \setlength{\parskip}{0pt}
  \setlength{\itemsep}{0pt plus 0.3ex}
}

\title{Exploring 3D-aware Lifespan Face Aging via Disentangled Shape-Texture Representations}
\name{Qianrui Teng \quad Rui Wang \quad Xing Cui \quad Peipei Li\sthanks{\ \ Corresponding author.} \quad Zhaofeng He}
\address{Beijing University of Posts and Telecommunications \\
\small \tt \{qrteng, wr\_bupt, cuixing, lipeipei, zhaofenghe\}@bupt.edu.cn}

\begin{document}\sloppy
\twocolumn[{%
\renewcommand\twocolumn[1][]{#1}%
\maketitle
\vspace{-10mm}
\begin{center}
    \includegraphics[width=0.7\linewidth]{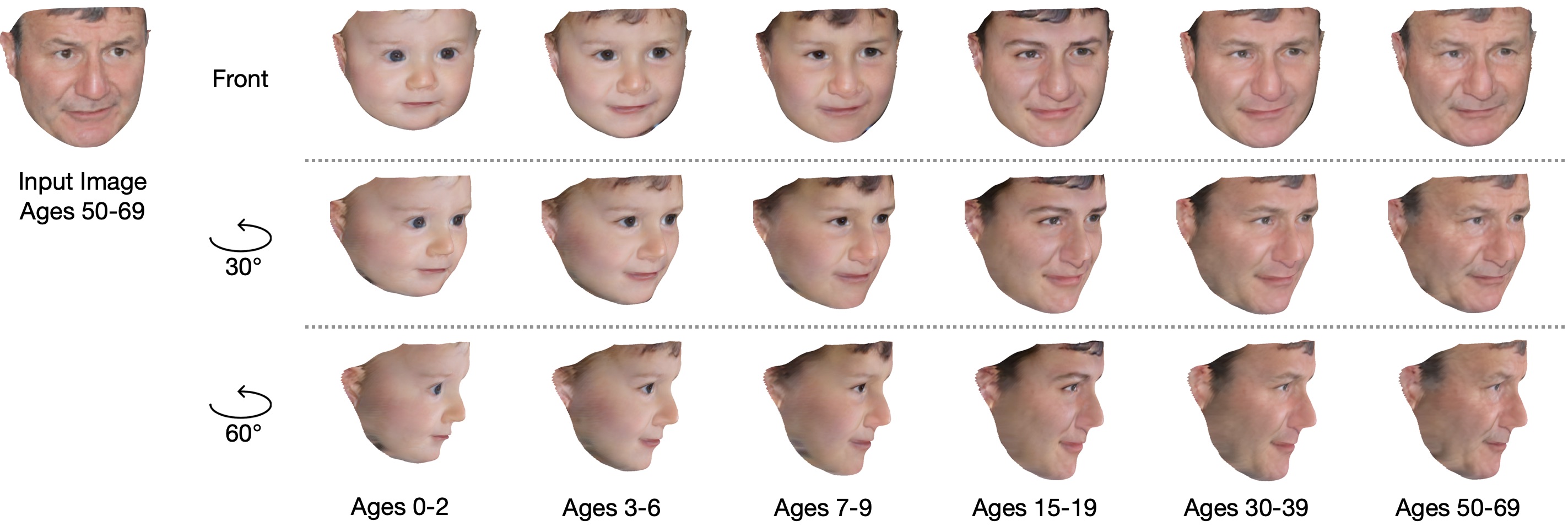}%
    \vspace{-3mm}
    \captionof{figure}{Given a 2D facial image from an arbitrary age group, our method enables producing lifespan 3D face aging results that demonstrate decent shape deformation and texture transformation quality.}
    \label{fig:3d-result}
    \vspace*{2mm}
\end{center}%
}]

\begin{abstract}
Existing face aging methods often focus on modeling either 
texture aging or using an entangled shape-texture representation to achieve face aging. However, shape and texture are two distinct factors that mutually affect the human face aging process. In this paper, we propose 3D-STD, a novel \textbf{3D}-aware\textbf{ S}hape-\textbf{T}exture \textbf{D}isentangled face aging network that explicitly disentangles the facial image into shape and texture representations using 3D face reconstruction. Additionally, to facilitate high-fidelity texture synthesis, we propose a novel texture generation method based on Empirical Mode Decomposition (EMD). Extensive qualitative and quantitative experiments show that our method achieves state-of-the-art performance in terms of shape and texture transformation. Moreover, our method supports producing plausible 3D face aging results, which is rarely accomplished by current methods.
\end{abstract}

\section{Introduction}
\label{sec:intro}
Face aging, as one of the most challenging tasks in face manipulation, aims at generating facial images depicting individuals at various ages across their entire lifespan, from infancy to old age. 
While numerous existing methods concentrate on modeling skin texture changes related to aging, such as wrinkles and furrows \cite{Wang_2018_CVPR,alalufOnlyMatterStyle2021,zossProductionReadyFaceReAging2022}, few address the crucial requirement of shape transformation in lifespan face synthesis.
Generally, human face aging process involves two main aspects: shape deformation and texture transformation \cite{FarkasAnthropometry}. Specifically, shape deformation primarily occurs during the youthful stages, while texture transformation persists throughout their entire lifespan. Both of these aspects are crucial manifestations of aging. 
While some studies have successfully addressed shape transformation in the context of lifespan face synthesis, they often failed to explicitly separate shape and texture changes, 
 leading to sub-optimal results  \cite{or-elLifespanAgeTransformation2020}.
He \textit{et al.} \cite{heDisentangledLifespanFace2021} highlight the importance of disentangling shape and texture factors in modeling the human aging process.
However, it is limited in its disentanglement capability as it operates solely within the constraints of the 2D image domain. 

In this paper, we propose a novel method that explicitly disentangles shape and texture features in a 3D space, resulting in completely disentangled representations for shape and texture. Subsequently, for both improved shape and texture transformation, we employ a dual-branch network to independently model the face aging process for the two factors. 

Meanwhile, accurately capturing the texture transformation pattern is indeed an essential part of the face aging task. A major challenge in facial texture aging lies in precisely modeling the high-frequency components of the texture. This may be partly due to an inherent limitation of deep generative models, known as spectral bias \cite{rahamanSpectralBiasNeural2019}, which indicates a degraded performance of generative networks when handling high-frequency contents. However, frequency information plays a vital role in face aging tasks. The loss of high-frequency information results in diminished facial texture aging quality, leading to blurred wrinkles, furrows, and other aging features. Therefore, we propose an innovative texture generation method by incorporating empirical mode decomposition (EMD) \cite{EMD}. EMD is employed to decompose the facial texture into distinct components with varying frequency ranges. By imposing constraints on each of these frequency ranges simultaneously, we can guide the model to overcome the spectral bias and generate high-quality textures.
To summarize, our contributions are three-fold:
\begin{itemize}[itemsep=0pt]
    \item We propose 3D-STD, a novel \textbf{3D}-aware\textbf{ S}hape-\textbf{T}exture \textbf{D}isentangled Face Aging Network. 3D-STD models face aging process in a disentangled way, resulting in superior transformations in both shape and texture.
    \item To enhance the texture generation quality, we propose a novel texture generation method based on empirical mode decomposition. This method helps the network capture subtle aging details, particularly in high-frequency ranges.
    \item Extensive qualitative and quantitative experiments show that our method achieves state-of-the-art performance. Notably, it is one of the earliest explorations to produce plausible 3D face aging results.
\end{itemize}

\section{Related work}
\textbf{Face Aging} is one of the most challenging subjects in facial image manipulation. It is an ill-posed problem and has to model extremely complex facial transformation process, where the shape and texture of faces show nonlinear changes with the increase of age.
In recent years, the applications of deep generative models like GANs and Diffusion Models have shown impressive performance in face aging tasks \cite{Wang_2018_CVPR,alalufOnlyMatterStyle2021,or-elLifespanAgeTransformation2020,heDisentangledLifespanFace2021,cusp,pada}.
Specifically, LATS \cite{or-elLifespanAgeTransformation2020} proposes a multi-domain image-to-image translation framework to perform face aging, which follows a StyleGAN-like generator architecture. It is able to synthesize smooth lifespan age translation results of a single identity.
DLFS \cite{heDisentangledLifespanFace2021} points out the importance of disentanglement in face aging problems initially, and proposes a disentangled aging framework that separates facial representations into identity, shape, and texture for modeling aging patterns.
Wu \textit{et al.} \cite{wuAdversarialUVTransformationTexture2022} use UV transformation to achieve facial texture aging.
\cite{heDisentangledLifespanFace2021,wuAdversarialUVTransformationTexture2022} are the works that most relevant to ours. Compared to \cite{heDisentangledLifespanFace2021}, we take a step further by explicitly disentangling shape and texture factors using 3D face reconstruction techniques. In \cite{wuAdversarialUVTransformationTexture2022}, only texture aging is achieved and it is unable to perform shape transformation. In contrast, our method conducts face aging fully in 3D space and can synthesize lifespan face aging results in 3D.

\textbf{Frequency-based Methods.} Some studies try to improve the texture aging quality through frequency-based methods. A3GAN \cite{A3GAN} utilized wavelet packet transform to capture texture features and enhance aging details by transforming the input images into the frequency domain.
AW-GAN \cite{AW-GAN} combines wavelet transform and attention mechanisms for aging on children and adults. Both \cite{A3GAN,AW-GAN} employ a wavelet-based multiscale discriminator to enhance aging details. Wavelet information is also used in \cite{global-local} in combination with the identity protection network. 
In contrast, in this paper, we explore EMD as the frequency-based method. It is a data-adaptive method that does not rely on fixed filters, giving larger flexibility and modeling capability in frequency space. 

\begin{figure*}
    \centering
    \includegraphics[width=0.8\linewidth]{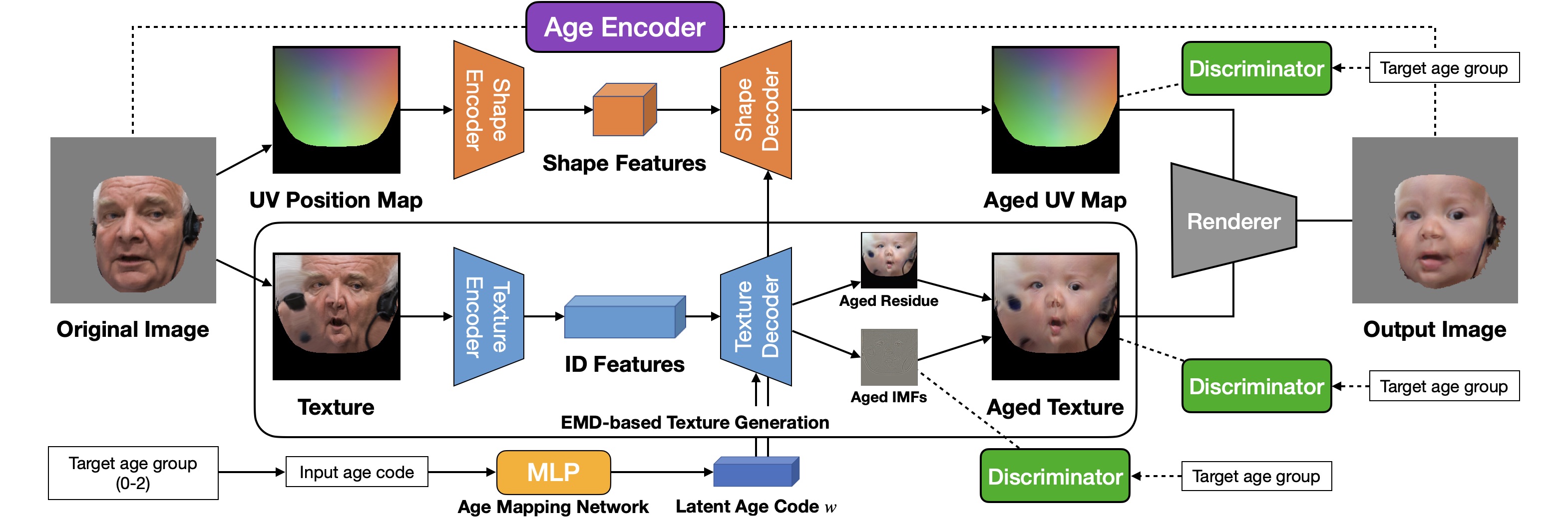}
    \vspace{-4mm}
    \caption{Overview of 3D-STD. The input is disentangled into 3D shape and texture representation, then dual-branch aging process is performed to get the aging result. EMD-based generation is used in texture branch to enhance texture aging quality.}
    \label{fig:network}
    \vspace{-3mm}
\end{figure*}

\section{Proposed method}
Our overall framework is shown in \cref{fig:network}. Given an input image $I_{ori}$ from source age group $src$ and a target age group $tgt$ ($src, tgt\in 1, ..., N$, $N$ is the total number of age groups), we want to transform $I_{ori}$ to $I_{tgt}$, which is the facial image in the desired age group $tgt$. First, we perform it by disentangling shape and texture information of the input image $I_{ori}$ into shape representation $P_{ori}$ and texture representation $T_{ori}$ (Sec.~\ref{3Dnetwork}), then we perform dual-branch face aging process to get $P_{tgt}$ and $T_{tgt}$. Specifically, to generate high-fidelity texture maps, we propose an EMD-based texture generation method (Sec.~\ref{sec:emd}). Finally, we employ a renderer to get the output $I_{tgt}$ from $P_{tgt}$ and $T_{tgt}$. 


\subsection{Preliminary}
\label{preliminary}
\textbf{Empirical Mode Decomposition (EMD)}. Originally proposed in \cite{EMD}, empirical mode decomposition is a signal processing algorithm used to analyze nonlinear and non-stationary data by obtaining local features and time-frequency distribution of the data \cite{bhuiyanFastAdaptiveBidimensional2008}. It is a multi-resolution technique that allows for the decomposition of signals into distinct, physiologically meaningful components, known as Intrinsic Mode Functions (IMFs). Each IMF is a narrow-band signal with a well-defined frequency range. Through an iterative decomposition process, a series of IMFs will be obtained in the order of high frequency to low frequency, and the sum of all IMFs can approximate the original signal.

The EMD algorithm was initially developed for analyzing one-dimensional (1D) signals, which has been extended to analyze 2D data/images, known as bi-dimensional EMD (BEMD). 
A 2D image $I$ can be decomposed using BEMD:
\begin{equation}
I=\sum_{i=1}^N C_i+R,   
\end{equation}
where $C_i, i=1, …, N$ denote the $i$th bi-dimensional IMF (BIMF) and $R$ is the trend within the data, also referred to as the last BIMF or residual.

\subsection{3D-aware shape-texture disentanglement}
\label{3Dnetwork}

Our network performs the aging process firstly by disentangling the shape and texture features of the input image. 
Given a 2D input, we use PRNet \cite{fengJoint3DFace2018} to predict the 3D face shape. We use UV position map \cite{fengJoint3DFace2018} and texture map to represent shape and texture information respectively.

UV position map is a 2D image recording 3D coordinates of all vertices in the face model. The UV position map can be expressed as $P(u_i,v_i)=(x_i,y_i,z_i)$, where $(u_i,v_i)$ represents the UV coordinate of $i$th vertex in the face model and $(x_i,y_i,z_i)$ represents the corresponding 3D coordinate of $i$th vertex. The coordinates of each vertex are predicted from the 2D input image by 3D reconstruction method.

From another perspective, we are also doing the normalization operation since we are mapping input faces with different locations and poses in 2D image space to a normalized UV space. In UV space, we get a canonical, pose-normalized view of a 3D face model. We believe it will bring great benefits in learning complex facial aging transformations and speed up the convergence of the network.
Once we get the UV position map, we can easily extract the facial texture $T$ from the input image, $T(u_i,v_i)=(R_i,G_i,B_i)$, which is the RGB value corresponding to each vertex of the face model.

We obtain $P_{ori}$ and $T_{ori}$ after the 3D-aware disentangling step. Subsequently, we perform the dual-branch aging.
For the shape branch, we obtain the shape features $f_s$ using a shape encoder $\ {E}_s$: $f_s=E_s(P_{ori})$. Then $f_s$ is decoded into aged position map $P_{gen}$ by a shape decoder $G_s$ conditioned on the latent age code $w$, expressed as:
\begin{equation}
P_{gen}=G_s(f_s, w).
\end{equation}
The latent age code $w$ is obtained by transforming the input age code $z$ through 8-layer MLPs, similar to \cite{karrasAnalyzingImprovingImage2020}. The input age code $z$ is a $50\times N$ element vector acquired identically as \cite{or-elLifespanAgeTransformation2020} according to the age group.

The texture branch resembles the shape branch, yet they differ from two critical points: 1) We assume that a person's identity information lies mostly in their texture information rather than the coarse shape information, thus we employ an identity encoder $E_{id}$ to extract the person's identity features: $f_{id}=E_{id}(T_{ori})$. 2) We employ a novel EMD-based texture generation method to output the aged texture map $T_{gen}$. The whole procedure can be expressed as:
\begin{equation}
T_{gen}=\mathcal{T}(f_{id},w),
\end{equation}
where $\mathcal{T}$ refers to the EMD-based texture generation module.
Finally, we employ a renderer to output our 3D aging result from a specific viewpoint. Note that if we render from the front view, we can get the traditional 2D aging result.

\subsection{EMD-based texture generation}
\label{sec:emd}

\begin{figure}[h]
    \centering
    \includegraphics[width=1\linewidth]{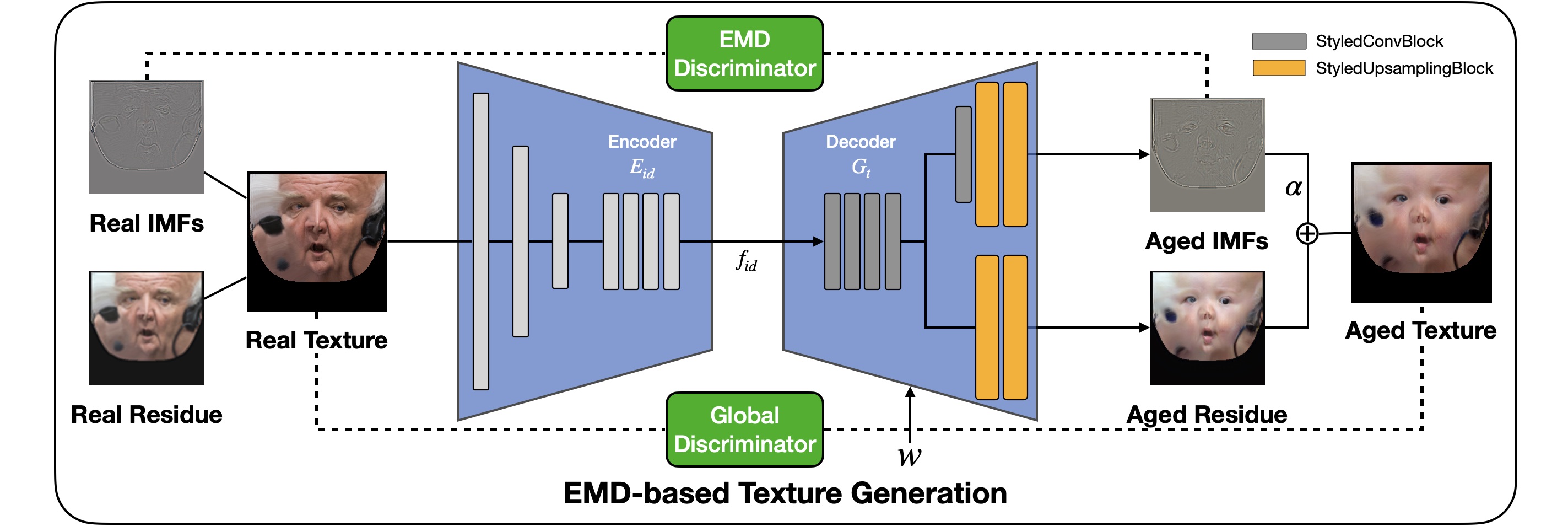}
    \vspace{-8mm}
    \caption{Illustration of EMD-based texture generation method.}
    \label{fig:emd}
    \vspace{-3mm}
\end{figure}

As mentioned in Sec.~\ref{sec:intro}, high-frequency content is a crucial aspect in face aging tasks since aging details manifest as high-frequency elements. Current generative networks have inherent limitations in modeling high-frequency content within the distribution. Therefore, in this section, we aim to address this challenge by incorporating EMD. Concretely, we mainly make three improvements as illustrated in \cref{fig:emd}: 1) To enhance the network's ability to learn high-frequency patterns, we force the generator to output the high-frequency components of the texture. 2) To further adapt the generator to this new generation paradigm, we re-designed the generator network architecture. 3) We also introduce additional losses to ensure the generation quality of high-frequency components.

As discussed in Sec.~\ref{preliminary}, a texture map $T$ can be decomposed by EMD algorithm into IMFs and the residue:
\begin{equation}
T={\Sigma_C}+R=\sum_{i=1}^N C_i+R,
\end{equation}
where $T,{\Sigma_C},R\in \mathbb{R}^{H\times W\times 3}$. Here, the total number $N$ of the IMFs is a hyperparameter and can be specified by user.
We note that the IMFs represent the top $N$ highest frequency elements of the input texture, while the residue image represents the overall trend, i.e., the low-frequency part.

After obtaining identity features $f_{id}$ as mentioned in Sec.~\ref{3Dnetwork}, we send it into the texture decoder $G_{t}$ and enforce it to output both the IMFs and the residue:
\begin{equation}
{\Sigma_C}_{gen},R_{gen}=G_{t}(f_{id},w_{}),
\end{equation}
where ${\Sigma_C}_{gen},R_{gen}\in \mathbb R^{H\times W\times 3}$.

We found that naively combining the generated IMFs and residue will cause training instability and result in network collapse. Thus, to reduce the difficulty of network training, we fuse the generated IMFs and the residue using a coefficient $\alpha$, thereby obtaining the final texture branch output:
\begin{equation}
T_{gen}=\alpha\cdot{\Sigma_C}_{gen}+R_{gen}.
\end{equation}

In terms of losses, we only constrain the final output $T_{gen}$ and the IMFs output ${\Sigma_C}_{gen}$.
We employ two discriminators to supervise the training of the texture branch, the global discriminator and EMD discriminator, which discriminate the output texture map $T_{gen}$ and the output IMFs ${\Sigma_C}_{gen}$ respectively.
For architecture improvement, we introduce an extra branch in the original decoder dedicated to generating high-frequency IMFs, while the original branch is used for generating the residue. The IMF branch takes in high-level semantic features from the original branch before upsampling.

\begin{figure*}[h]
    \centering
    \includegraphics[width=0.90\linewidth]{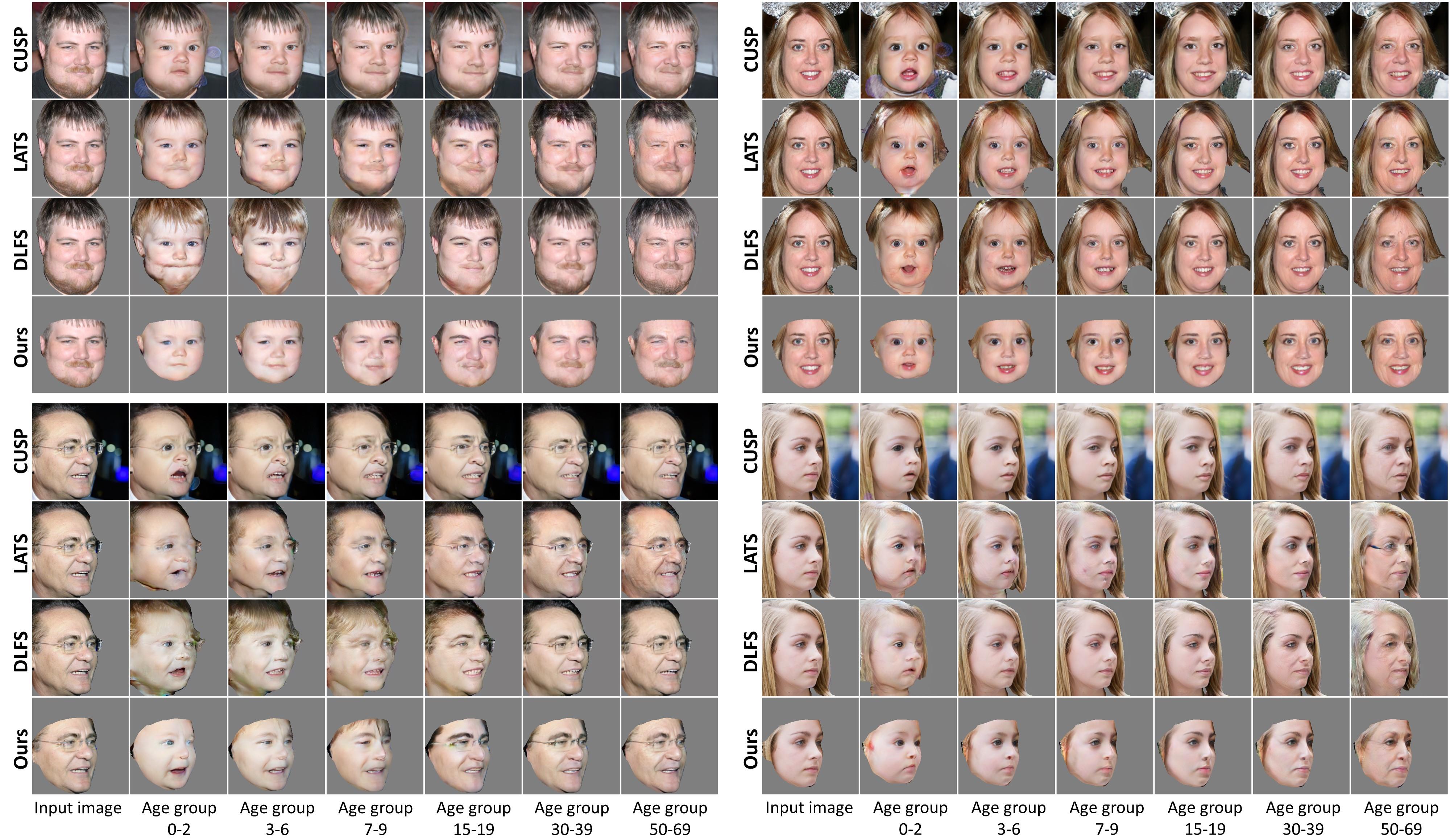}
    \vspace{-5mm}
    \caption{Qualitative comparison with three state-of-the-art methods on FFHQ-Aging test set.}
    \label{fig:result_img}
    \vspace{-3mm}
\end{figure*}

\subsection{Training losses}
\label{sec:loss}
We utilized five types of losses to train the network.

\noindent\textbf{Adversarial loss.} We use the conditional adversarial loss based on the source and target age group:
\begin{equation}
\mathcal L_{adv}=E_{x}[\log{D_{}(x|src)}]+E_{x,tgt}[\log{(1-D_{}(y_{tgt}|tgt))}],
\end{equation}
where $x$ means input image from $src$ age group, $y_{tgt}$ is the generated image conditioned on the target group $tgt$.

\noindent\textbf{Reconstruction loss.} When conditioned on the source age group, the output image $y_{src}$ should be exactly the same as $x$:
\begin{equation}
\mathcal L_{rec}={\|y_{src}-x\|}_1.    
\end{equation} 

\noindent\textbf{Cycle loss.} To preserve the identity throughout the age translation, cycle loss \cite{zhuUnpairedImageToImageTranslation2017} is introduced as:
\begin{equation}
\mathcal L_{cyc}={\|y_{cyc}-x\|}_1,   
\end{equation}
where $y_{cyc}=G(E(y_{tgt}),w_{src})$.

\noindent\textbf{Identity loss.} We further preserve the identity information by minimizing the difference of the outputs of the identity encoder through the translation process:
\begin{equation}
\mathcal L_{id}={\|E_{id}(x)-E_{id}(y_{tgt})\|}_1.    
\end{equation}

\noindent\textbf{Age loss.} To enhance aging accuracy, we minimize the difference between the output of an age encoder $E_{age}$ and the input age code for both real and generated images:
\begin{equation}
\mathcal L_{age}={\|E_{age}(y_{tgt})-z_{tgt}\|}_1+{\|E_{age}(x)-z_{src}\|}_1.
\end{equation}

For the shape branch, we utilize $\mathcal L_{rec},\mathcal L_{cyc}$ and $\mathcal L_{adv}$. As stated in Sec.~\ref{3Dnetwork}, our assumption is that human face shape changes drastically during the stage from infancy to adulthood and it is infeasible to acquire identity-related features from it. The loss of the shape branch is defined as follows:
\begin{equation}
\mathcal L^s=\mathcal L_{rec}^s+\mathcal L_{cyc}^s+\mathcal L_{adv}^s.
\end{equation}
Here, $y$ and $x$ in each loss definition refer to position map $P$.

For the texture branch, we utilize all five types of losses:
\begin{equation}
\mathcal L^t=\mathcal L_{rec}^t+\mathcal L_{cyc}^t+\mathcal L_{adv}^t+\mathcal L_{age}^t+\mathcal L_{id}^t.
\end{equation}
Here, $y$ and $x$ in each loss definition refer to texture map $T$.

Moreover, We refactor $\mathcal L_{rec}^t,\mathcal L_{cyc}^t$ and $\mathcal L_{adv}^t$ along with the incorporation of the EMD-based texture generation process. When performing the losses, we get the ground truth IMFs ${\Sigma_C}^{ori}$ by performing EMD on the original texture $T_{ori}$. Each refactored loss adds an IMF term compared with before, e.g., ${\mathcal L_{rec}^t}'=\mathcal L_{rec}^t+\mathcal L_{rec}^{IMF}$.
In the IMF term of each loss, $y$ and $x$ refer to the IMF component ${\Sigma_C}$.
The refactored loss for the texture branch is:
\begin{equation}
\mathcal L^t={\mathcal L_{rec}^t}'+{\mathcal L_{cyc}^t}'+{\mathcal L_{adv}^t}'+\mathcal L_{age}^t+\mathcal L_{id}^t.
\end{equation}
The total loss is obtained by combining losses of two separate branches, with a hyperparameter $\lambda_s$ to balance the weight:
\begin{equation}
\mathcal L=\lambda_s\cdot \mathcal L^s+\mathcal L^t.    
\end{equation}

\section{Experiments}
\label{sec:exp}

\noindent\textbf{Datasets.} We use pruned FFHQ-Aging dataset \cite{or-elLifespanAgeTransformation2020} to train our model. It contains 70000 images labeled in 10 age groups.
It is then pruned to 28701 images dedicated for lifespan face synthesis. The pruned dataset is divided into 6 discrete age groups: \{0-2, 3-6, 7-9, 15-19, 30-39, 50-69\}, which has 14232 training images and 198 testing images for male, and 14066 training images and 205 testing images for female. 

\noindent\textbf{Evaluation metrics.} 
To evaluate the aging accuracy, we employ the age estimator of \cite{hsuAgeTransGANFacialAge2022} to estimate the age mean absolute error (Age MAE), and we provide additional results using Face++ API in Sec.~\ref{sec:add_quant}. We employ the pre-trained ArcFace \cite{arcface} for identity preservation evaluation. 
For human evaluation, we ask volunteers to compare the performance with different aging methods. The results can be seen in Sec.~\ref{sec:add_quant}.

\subsection{Comparison with Face Aging Methods}
We compare our method with three state-of-the-art aging methods, including LATS \cite{or-elLifespanAgeTransformation2020}, DLFS \cite{heDisentangledLifespanFace2021}, and CUSP \cite{cusp}.

\noindent \textbf{Qualitative comparison.} Fig.~\ref{fig:result_img} shows a qualitative comparison with the state-of-the-art evaluated on the FFHQ-Aging dataset, where we transform each input image into a total of 6 age groups. 
Compared to other methods, ours has a slightly narrower visible region. This is because the employment of 3D face reconstruction has removed all unrelated parts in the image, e.g., background, hair, etc. It does not harm the main goal since our method has preserved the entire face region. 
From the results, it can be clearly observed that our method has achieved superior shape transformation performance, especially for early age stages when large shape deformation takes part. LATS and DLFS's shape transformation is not sufficient enough for fitting in the younger age groups (e.g. 0-2, 3-6). CUSP shows little variation in face shape for the entire aging process. 
It also can be seen that our method supports producing purer textures (bottom left), while the performance of the other three may deteriorate with the existence of accessories like glasses. Furthermore, our method performs well on faces with extreme poses (bottom right), where LATS and DLFS fail to synthesize high-quality results and CUSP loses the original identity after transformation. This can be credited to our canonical facial representation, where shape and texture information is stored as normalized maps.

\begin{table}[h]
    \vspace{-3mm}
    \caption{Quantitative comparison on FFHQ-Aging test set.}
    \centering
    \begin{adjustbox}{max width=\linewidth}

        \resizebox{\linewidth}{!}{%
        \begin{tabular}{ccccccccc} \toprule
             \multirow{2}{*}{Method}&  \multicolumn{7}{c}{Aging Accuracy ($\downarrow$)} &Identity\\ \cmidrule(lr){2-8}
             &  0-2&  3-6&  7-9&  15-19&  30-39& 50-69 &Overall & Preservation ($\downarrow$) \\ \midrule
             LATS&  2.56&  2.94&  4.92&  8.77&  \textbf{0.47}& \underline{3.00}&3.78&\underline{0.52}
    \\ 
             DLFS&  1.47&  2.59&  3.49&  6.76&  0.93&  5.05&3.38&\textbf{0.49}
    \\  
     CUSP& \textbf{0.03}& \underline{0.12}& \underline{1.27}& \textbf{0.96}& 3.39& 7.53&\underline{2.21}&0.63
    \\ 
             Ours&  \underline{0.29}&  \textbf{0.05}&  \textbf{0.91}&  \underline{6.60}&  \underline{0.76}& \textbf{2.61}&\textbf{1.87}&0.53\\ \bottomrule
        \end{tabular}
        }
    \end{adjustbox}
      \vspace{-2mm}

    \label{tab:quantitative}
\end{table}

\noindent \textbf{Quantitative comparison.} \cref{tab:quantitative} shows the quantitative comparison of our method with the state-of-the-art. We can see our method achieves the best performance on aging accuracy. Our method has surpassed our baseline (LATS) by a large margin. It also can be seen that our method performs particularly well in young age groups, which is consistent with our shape-texture disentangling approach. Our method also performs best in the last age group (50-69), which matches our EMD-based texture generation approach for better representation capacity on aging patterns like wrinkles and furrows.
The identity preservation result shows that DLFS performs the best on this metric, while LATS and our method show slightly worse performance.  CUSP's aging accuracy results are closest to ours, but it fails to preserve people's identities at a high level during the transformation.
However, it is worth noting that existing facial recognition systems often do not perform well in cross-age face recognition tasks \cite{huangWhenAgeInvariantFace2023}. It means that evaluation results for the same identity may vary when there is a large age gap. Thus, aging accuracy and identity preservation may conflict when encountering radical age transformations.

\subsection{Ablation Study}
\begin{figure}[h]
    \centering
    \includegraphics[width=0.8\linewidth]{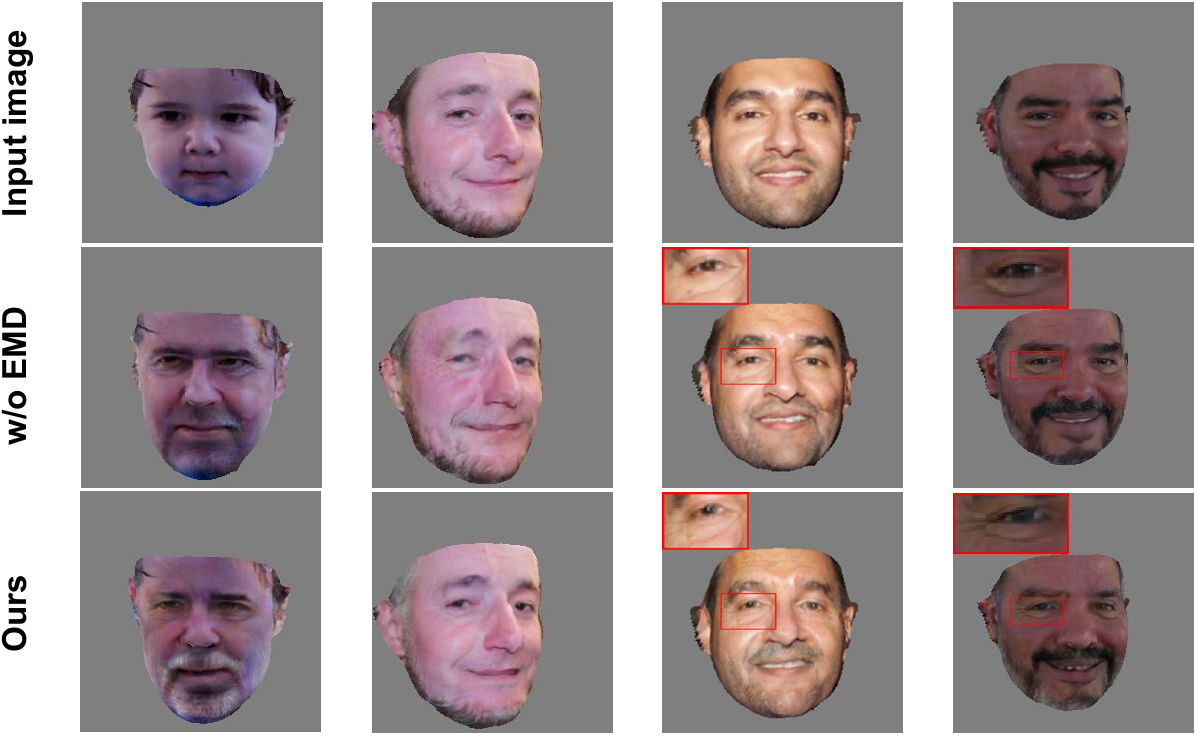}
    \vspace{-3mm}
    \caption{Ablation study of 3D-STD. 
    }
    \label{fig:emd_ablation}
    \vspace{-5mm}
\end{figure}

\textbf{The effectiveness of EMD-based texture generation.} We train a model without EMD-based texture generation method to validate its effectiveness. Since the high-frequency aging patterns are most pronounced in elder age groups, we make a comparison by transforming the input into the last age group (50-69). From Fig.~\ref{fig:emd_ablation}, we see the EMD-based texture generation can learn the aging pattern of hairs (e.g., beards and hair on the temples) more effectively, and strengthen the details around the eyes, like crow's feet and melanin deposition (third column around the eyes).
\cref{tab:ablation} proves that the incorporation of EMD-based texture generation will further improve the aging accuracy of our method.
We also analyze in frequency domain to validate its effectiveness in reducing the distribution gap.
We calculate the mean squared difference of the magnitude of the Fourier transform results between the generated images and the real images.
The result shows that the generated images without EMD-based texture generation have a larger spectral difference. It indicates that the introduction of EMD-based generation does benefit the network in terms of reducing the distribution gap between real and generated images in the frequency domain.



\begin{table}[h]
    \vspace{-3mm}
    \caption{Ablation study of 3D-STD.}
    \centering
    \setlength{\tabcolsep}{6mm}
    \resizebox{\linewidth}{!}{%
    \begin{tabular}{lccc} \toprule
          &Age group&  W/o EMD& Ours\\ \midrule
          \multirow{4}{*}{Aging Accuracy($\downarrow$)}&3-6&  0.7& \textbf{0.1}\\
          &30-39&  \textbf{0.2}& 0.6\\
          &50-69&  5.5& \textbf{2.2}\\
          &\textbf{Overall}&  2.0& \textbf{1.9}\\ \midrule
          \multicolumn{2}{c}{Spectral Difference($\downarrow$)}&  6.82& \textbf{5.76}\\ \bottomrule
    \end{tabular}
    }

    \vspace{-2mm}
    \label{tab:ablation}
\end{table}

\noindent \textbf{3D aging results.} As shown in Fig.~\ref{fig:3d-result}, our method innovatively supports producing plausible 3D face aging results along lifespan longitude. We present more of our 3D face aging results in Sec.~\ref{sec:add_3D} of supplementary material.

\section{Conclusion}
We present a novel 3D-aware shape-texture disentangled approach to achieve high-quality shape and texture aging. To achieve the disentanglement, we utilize 3D face reconstruction to get canonical shape and texture representations. 
To enable high-quality texture aging, we propose an EMD-based texture generation approach for compensating high-frequency details. Extensive experiments have demonstrated the effectiveness of our method. We hope that our work can advance the research progress in 3D face aging, which is an unexplored area that holds significant research value.

\bibliographystyle{IEEEbib}
\bibliography{icme2023template}
\clearpage

\FloatBarrier


\onecolumn
\appendix

\vspace{-2cm}
\begin{center}
\LARGE\textbf{Appendix}
\end{center}

\FloatBarrier

\section{Implementation details}
We use our own implementation of the FBEMD algorithm \cite{bhuiyanFastAdaptiveBidimensional2008} to conduct the EMD process in our framework. We first normalize the texture map to [-1, 1] range and decompose it into four EMD components, i.e., three IMFs and one residue. 
Our model is implemented in PyTorch. Each input image is resized to 256×256. Following \cite{or-elLifespanAgeTransformation2020}, we train male’s model and female’s model separately. Each model is trained for 80 epochs on a single RTX 3090 using batch sizes of 2. We use the Adam optimizer with $\beta_1=0,\beta_2=0.999$, the learning rate of 0.001. And the learning rate is decayed by 0.5 after 50 and 100 epochs. For female model, we use the learning rate warm-up in the first 5 epochs, where the learning rate increases linearly to the set value. For hyperparameters, $\lambda_{rec}, \lambda_{cyc}, \lambda_{id}, \lambda_{age}$ are used in the loss function to balance the weight of respective losses, and $\lambda_{emd}$ is used in refactored loss of texture branch to adjust the weight of $\mathcal{L}^{IMF}$. $\lambda_{s}$ is used to balance the weight of two branches as stated in the paper. They are set to $\lambda_{rec}=10,\lambda_{cyc}=10,\lambda_{id}=1,\lambda_{age}=1,\lambda_{emd}=0.3,\lambda_{s}=0.3$. 

For the evaluation of age MAE in Sec. \ref{sec:exp}, following \cite{hsuAgeTransGANFacialAge2022}, we employ a rectified age estimator to estimate the age of result images, which is more appropriate for lifespan age synthesis since it performs more consistently across all ages, especially more accurate for estimating ages of infants and children, the results are shown in Tab. \ref{tab:quantitative} of main paper. In the ablation study, we use the following approach to obtain the spectral difference result: we first gather all generated images and sample 5000 images randomly in the original dataset, and calculate their average spectrum using Fourier transform $\mathbb E[\mathcal{F}(x)]$. Then we calculate the mean squared difference of the magnitude of the Fourier transform results between the generated images and the real images, i.e., $\frac{1}{XY}\sum_{x,y}(\|\mathbb E[\mathcal{F}(gen)]\|-\|\mathbb E[\mathcal{F}(real)]\|)^2$.

\section{Additional qualitative comparisons}
\begin{figure}
    \centering
    \includegraphics[width=1\linewidth]{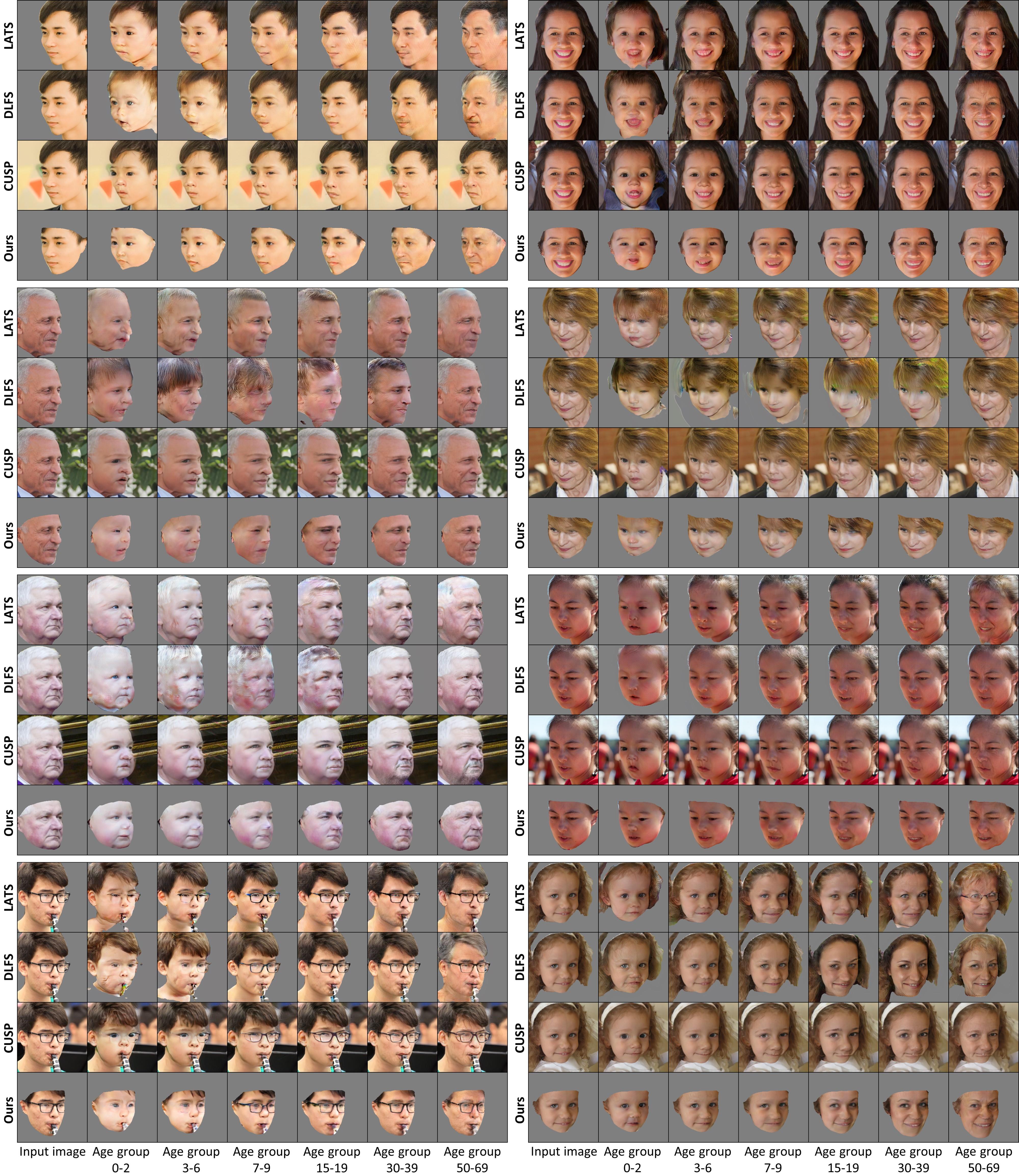}
    \caption{Qualitative comparisons of 3D-STD with LATS, DLFS and CUSP.}
    \label{fig:supp_1_1}
\end{figure}

\begin{figure}
    \centering
    \includegraphics[width=1\linewidth]{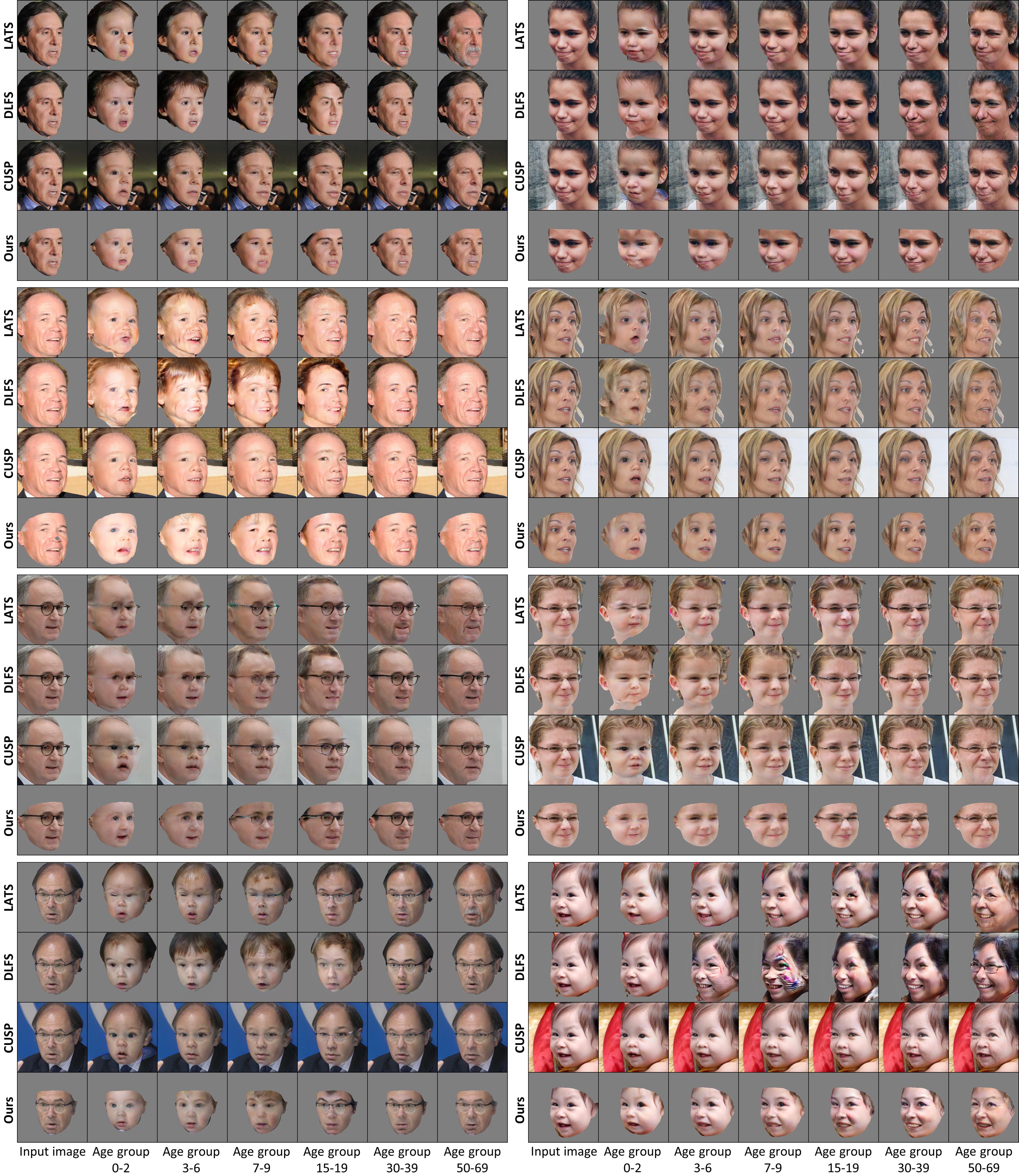}
    \caption{Qualitative comparisons of 3D-STD with LATS, DLFS and CUSP.}
    \label{fig:supp_1_2}
\end{figure}

We present more of the qualitative comparisons of our method with LATS \cite{or-elLifespanAgeTransformation2020}, DLFS \cite{heDisentangledLifespanFace2021} and CUSP \cite{cusp} in Figs. \ref{fig:supp_1_1} and \ref{fig:supp_1_2}.

\section{3D face aging results}
\label{sec:add_3D}

\begin{figure}
    \centering
    \includegraphics[width=1\linewidth]{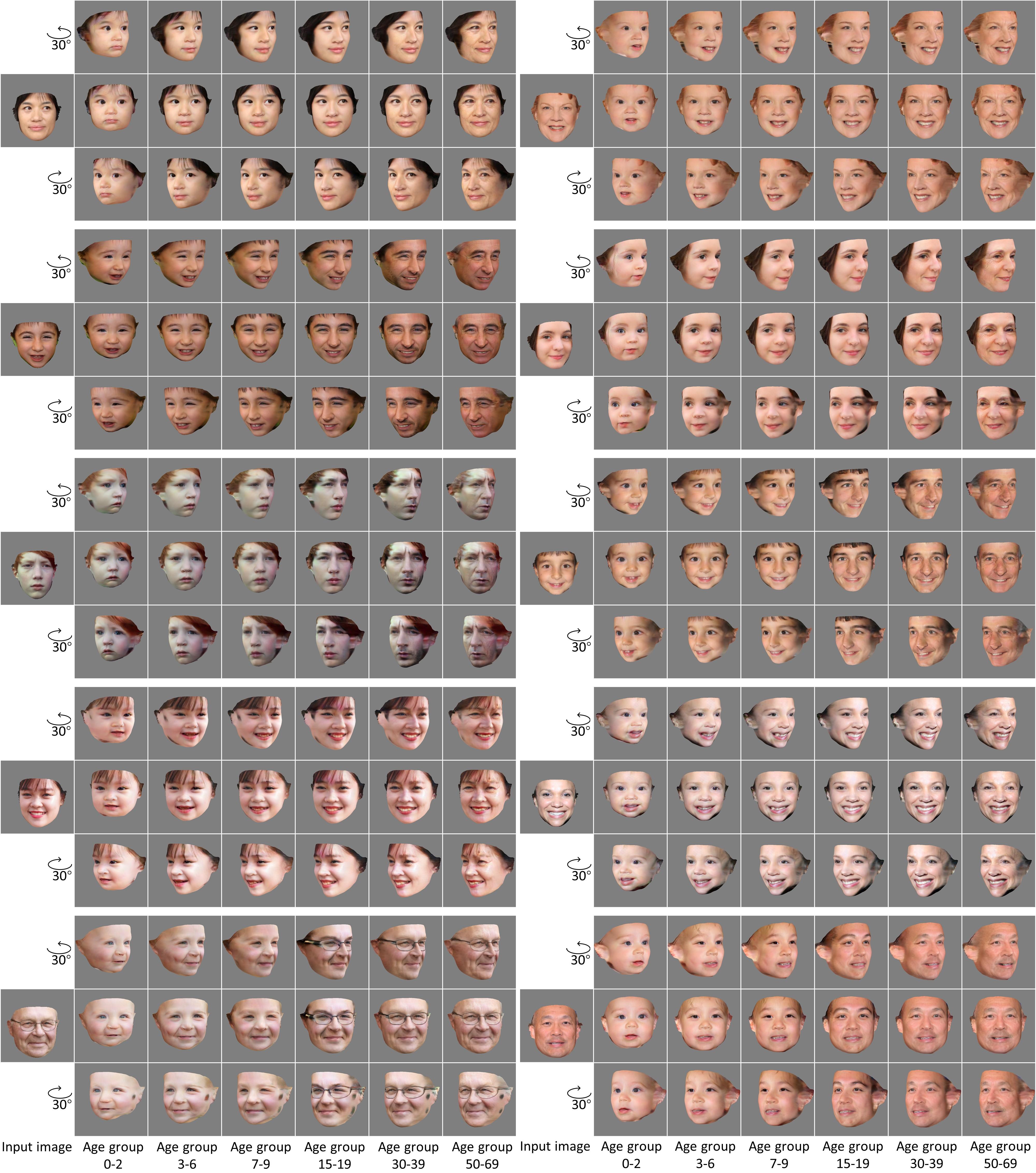}
    \caption{3D face aging results of 3D-STD shown from three perspectives.}
    \label{fig:supp_2_1}
\end{figure}

\begin{figure}
    \centering
    \includegraphics[width=1\linewidth]{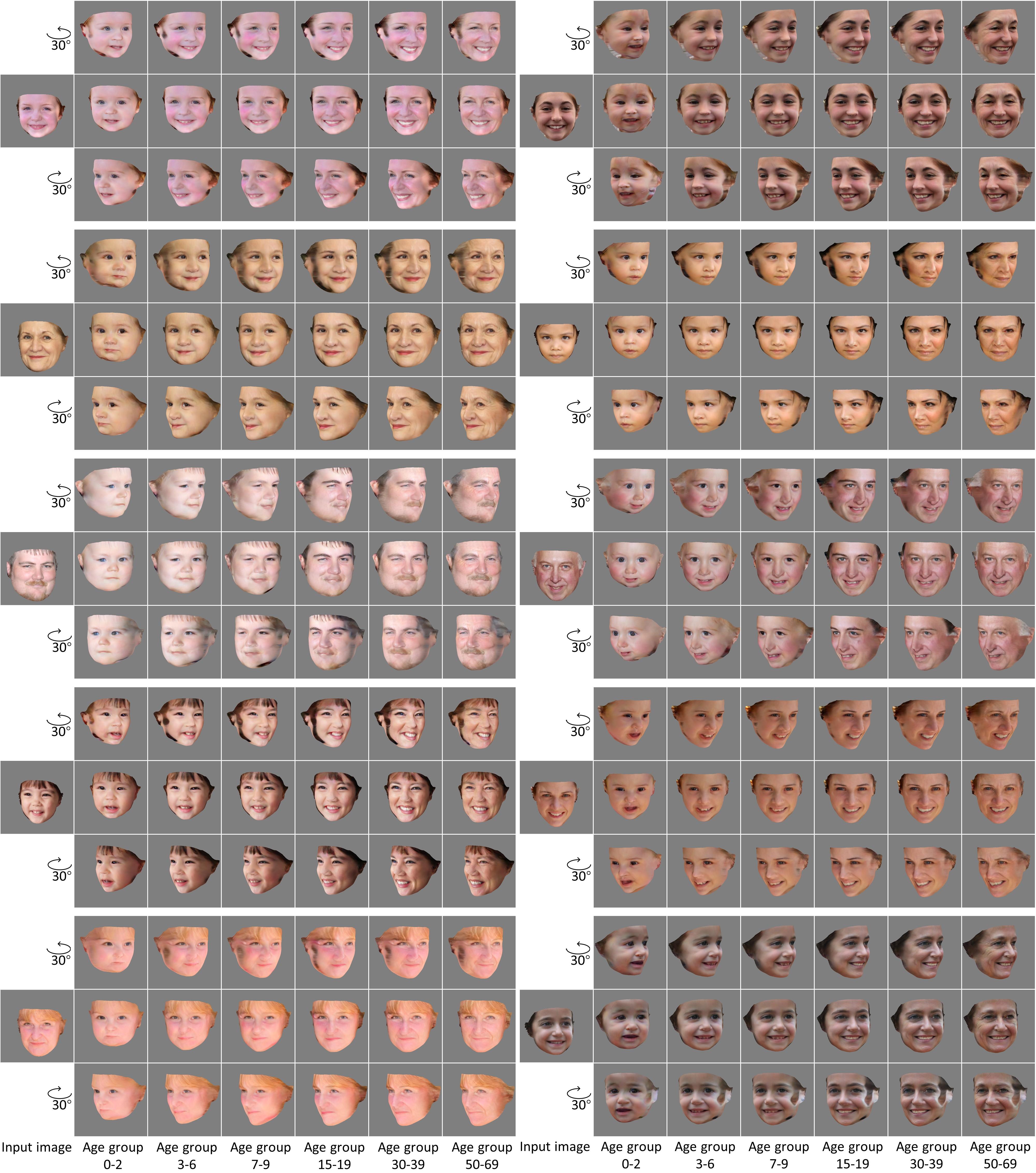}
    \caption{3D face aging results of 3D-STD shown from three perspectives.}
    \label{fig:supp_2_2}
\end{figure}

\begin{figure}
    \centering
    \includegraphics[width=1\linewidth]{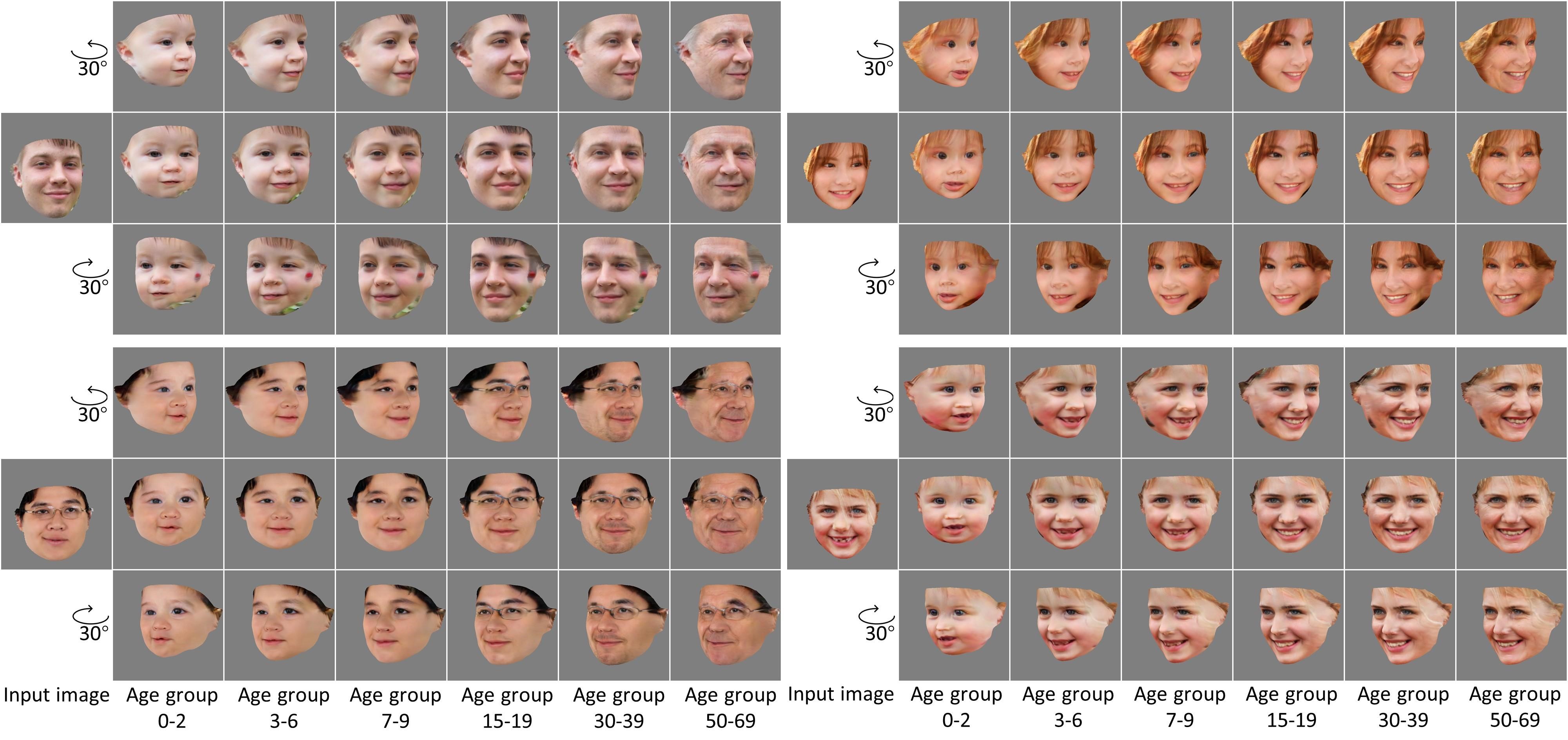}
    \caption{3D face aging results of 3D-STD shown from three perspectives.}
    \label{fig:supp_2_3}
\end{figure}

In this section, we present more of our novel 3D face aging results in Figs. \ref{fig:supp_2_1} - \ref{fig:supp_2_3}. Each face aging result is shown from three perspectives: the front view, rotated 30 degrees to the left and rotated 30 degrees to the right. Note that our 3D aging results are stored as 3D models so that they can be viewed from arbitrary angles apart from the above.

\section{Additional quantitative results}
\label{sec:add_quant}
\begin{table}
    \centering

    \caption{Human evaluation results on FFHQ-Aging. }
    \begin{tabular}{lccccc} \toprule
          &Age group&  LATS& DLFS&CUSP& \textbf{Ours}\\ \midrule
          \multirow{4}{*}{Aging Accuracy($\uparrow$)}&3-6&  2.47& \underline{2.67}& 2.22& \textbf{3.72}\\
          &30-39&\textbf{3.78}& 3.48& 1.76& \underline{3.58}\\
          &50-69&  \underline{3.66}     & 3.42     & 2.26     & \textbf{4.16}     \\
          &\textbf{Overall}&  \underline{3.30}& 3.19& 2.08& \textbf{3.82}\\ \midrule
          \multirow{4}{*}{ID Preservation($\uparrow$)}&3-6&  \textbf{3.02}& 2.95     & 2.92& \underline{2.97}\\
          &30-39&  \textbf{3.51}& 2.91& \underline{3.26}& 2.63\\
          &50-69&  3.06     & 2.74     & \textbf{3.68}     & \underline{3.44}     \\
          &\textbf{Overall}&  \underline{3.19}& 2.86& \textbf{3.28}& 3.01\\ \midrule
          \multirow{4}{*}{Aging Quality($\uparrow$)}&3-6&  2.70& \underline{3.02}& 2.45     & \textbf{3.55}     \\
          &30-39&  \textbf{3.52}& 3.30& 2.61& \underline{3.37}\\
          &50-69&  2.98     & \underline{3.14}     & 2.50& \textbf{3.96}     \\
          &\textbf{Overall}&  3.06& \underline{3.15}& 2.52& \textbf{3.62}\\ \bottomrule
    \end{tabular}

    \label{tab:human_evaluation}
\end{table}
\begin{table}
    \centering

    \caption{Quantitative comparison of age MAE using Face++ API.}
        \begin{tabular}{cccccccc} \toprule
             \multirow{2}{*}{Method}&  \multicolumn{7}{c}{Aging Accuracy ($\downarrow$)} \\ \cmidrule(lr){2-8}
             &  0-2&  3-6&  7-9&  15-19&  30-39& 50-69 &Overall \\ \midrule
             LATS&  11.5&  17.0&  20.9&  11.7&  6.1& 5.1&12.0\\ 
             DLFS&  11.0&  16.8&  18.5&  10.4&  5.8&  2.2&10.8\\  
     CUSP& 8.8& 14.9& 19.3& 10.5& \textbf{5.5}& \textbf{0.8}&10.0\\ 
             Ours&  \textbf{7.1}&  \textbf{14.2}&  \textbf{15.0}&  \textbf{8.8}&  6.3& 4.8&\textbf{9.4}\\ \bottomrule
        \end{tabular}

    \label{tab:face++_quantitative}
\end{table}
\begin{figure}
    \centering
    \includegraphics[width=1\linewidth]{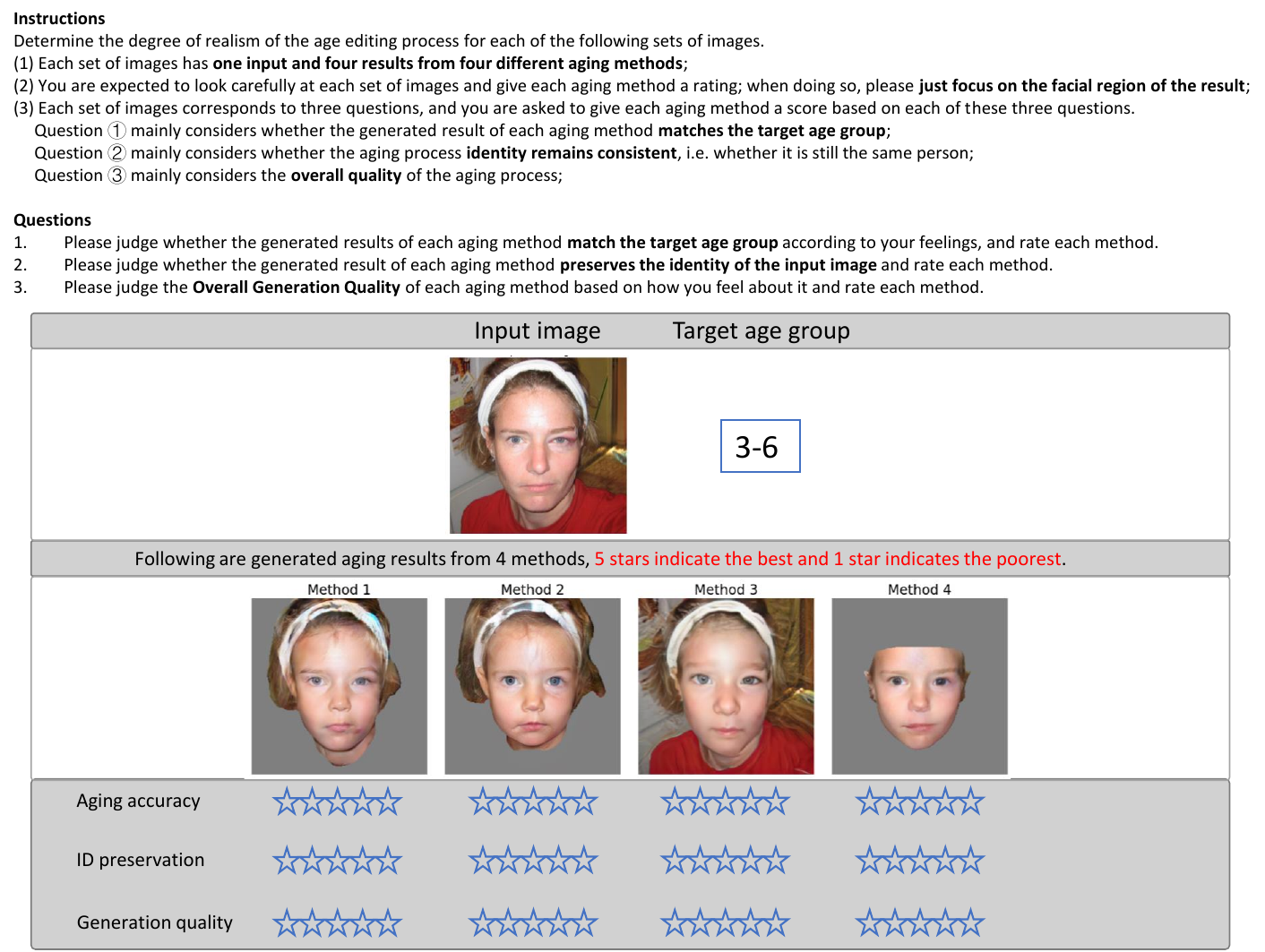}
    \caption{User study interface.}
    \label{fig:US_interface}
\end{figure}
In this section, we provide more of the quantitative results and comparisons of our method. 

\noindent\textbf{Age MAE by Face++ API.} In Tab. \ref{tab:face++_quantitative}, we provide the age MAE evaluation result using the common Face++\footnote{Face++ Face detection API: https://www.faceplusplus.com/}. It can be learned from the table that the results obtained using Face++ API are consistent with the results in the main paper shown in Tab. \ref{tab:quantitative}, where our method consistently outperforms the other three methods.

\noindent \textbf{User study.} For conducting the user study, we ask volunteers to compare the performance with different aging methods. Following SAM and PADA \cite{alalufOnlyMatterStyle2021,pada}, we experiment on three distinct age groups (3-6, 30-39, 50-69), representing the age phases of toddler, adult and the elderly respectively. Specifically, we invite 30 volunteers and ask them to score the results of different methods, according to aging accuracy, identity preservation and overall aging quality. The interface when conducting the user study is shown in Fig. \ref{fig:US_interface}. Each volunteer is randomly allocated 30 reference image sets. We provide the human evaluation results in Tab. \ref{tab:human_evaluation}. 
The result shows our method consistently outperforms the other three methods in aging accuracy and aging quality. 
For ID preservation, we hypothesize that the radical shape transformation brought by our method may cause the interviewees to prefer other methods over ours. However, we argue that the face aging results including shape deformation are more natural and realistic compared to those without shape deformation. This part of the changes does not include identity-related information.

\end{document}